# Automated Discontinuity Set Characterisation in Enclosed Rock Face Point Clouds Using Single-Shot Filtering and Cyclic Orientation Transformation

Dibyayan Patra[1], Pasindu Ranasinghe[1], Bikram Banerjee[2], Simit Raval[1,*]

[1]School of Minerals and Energy Resources Engineering, University of New South Wales, Sydney, NSW, Australia.

[1]School of Surveying and Built Environment, University of Southern Queensland, Toowoomba, QLD, Australia.

[*]Corresponding Author Email: simit@unsw.edu.au

**Abstract:** Characterisation of structural discontinuity sets in exposed rock faces of underground mine cavities and tunnels is essential for assessing rock-mass stability, excavation safety, operational efficiency, and mine design. UAV and other mobile laser-scanning techniques provide efficient means of collecting point clouds from rock faces. However, the development of a robust and efficient approach for automatic characterisation of discontinuity sets in real-world scenarios, such as fully enclosed rock faces in mine cavities, remains an open research problem. In this study, a new approach is proposed for fully automatic discontinuity set characterisation that uses a single-shot filtering strategy alongside an innovative cyclic orientation transformation scheme and a hierarchical clustering technique. The single-shot filtering step isolates planar regions while robustly suppressing noise and high-curvature artefacts in one pass using a signal-processing technique. To address the limitations of Cartesian clustering on polar orientation data, a cyclic orientation transformation scheme is developed, enabling accurate representation of dip angle and dip direction in Cartesian space. The transformed orientations are then characterised into discontinuity sets using a hierarchical clustering technique, which handles varying density distributions and identifies clusters without requiring user-defined set numbers. The proposed approach is evaluated on real-world enclosed rock-face data acquired from mine stopes. The accuracy of the method is validated against ground truth obtained using manually handpicked discontinuity planes identified with the Virtual Compass tool, as well as widely used automated structure mapping techniques. The proposed approach outperforms the other techniques by exhibiting the lowest mean absolute error in estimating discontinuity set orientations in real-world stope data with errors of ±1.95° and ±2.20° in nominal dip angle and dip direction, respectively, and dispersion errors lying below ±3°.

**Keywords:** Discontinuity characterisation, structure mapping, UAV laser scanning, 3D point cloud, rock mass characterisation, underground mine.

## 1. Introduction

Rock-mass discontinuities, also known as geological or structural discontinuities, are features such as joints, fractures, bedding planes, and cleavage that arise from changes in the physical or mechanical properties of rock induced by geological deformation processes in the Earth's crust. These mechanical breaks exhibit very low tensile strength and show up on exposed rock surfaces as distinct planes of weakness during blasting, development, and excavation activities in underground mining and civil engineering operations. A detailed understanding of discontinuity plane orientation, spacing, persistence, and mechanical properties is essential for analysing rock-mass stability, which in turn underpins excavation safety, support structure design, ore-recovery optimisation, and broader mine and civil-design tasks [1-4]. The advent of 3D data acquisition technologies, particularly UAV and other mobile laser-scanning systems, has transformed structural mapping by enabling rapid and accurate capture of rock-face point clouds for discontinuity analysis [2, 5]. Characterising individual discontinuity planes into sets of similar orientations is a crucial step in rock-mass assessment, as discontinuities often occur in parallel, spatially separated groups whose members typically tend to share comparable geometric and mechanical properties [6, 7]. Identifying and analysing these discontinuity sets therefore provides a more coherent and efficient basis for rock-mass characterisation and stability evaluation.

Rock mass discontinuity set characterisation techniques in 3D point cloud can be divided into three primary types based on their level of automation: manual, semi-automatic and automatic. Manual methods for discontinuity mapping involve direct user interaction to identify discontinuities and determine their orientation. This process requires the user to manually select points that correspond to discontinuity planes by inspecting the 3D point cloud visually. Lastly, the orientation of those planes can then be estimated using techniques like plane fitting, in which a virtual compass is placed over the selected points to determine the dip angle and dip direction of the discontinuity plane [8, 9]. Semi-automated techniques, on the other hand, aim to streamline the process of discontinuity mapping by incorporating user-defined parameters for discontinuity identification. Thresholds for angles, lengths, or deviations from fitted planes are a few of such parameters. Semi-automated approaches, while still allowing user inputs, aim for a balance between efficiency and accuracy by automating certain aspects of the discontinuity identification process. However, they are still prone to subjectivity and potential biases due to their significant reliance on human judgment [2, 10]. A few of the significant algorithms and software that fall under this category and are employed in mining and civil applications are Maptek PointStudio [11], Discontinuity Set Extractor [12-14], DiAna [15] and CloudCompare facet plugin [16]. Contrary to the manual and semi-automated methods, automated approaches perform discontinuity mapping without the requirement of any user input. Such algorithms, based on the data characteristics, either predefine the required parameters or adjust them adaptively. Such algorithms are also capable of processing large datasets effectively by leveraging computational techniques like clustering and unsupervised learning, but their degree of effectiveness heavily depends on the data quality and algorithmic robustness [1, 17].

Many of the automated discontinuity mapping algorithms utilise clustering techniques on normal vectors of a point cloud. Such clustering techniques group together points into discontinuities based on similar characteristics like orientation, identified from the normal vectors. The most used clustering algorithms in discontinuity mapping are k-means [18, 19], fuzzy k-means [20, 21], DBSCAN [12], ISODATA [22], and clustering by fast search and find of density peaks [23]. These techniques can identify discontinuities effectively in homogeneous datasets, but struggle with noisy real-world point clouds due to the significant variations in point normal vectors resulting from the noise. Region growing methods for identifying discontinuities are another type of algorithm in which variations between normal vectors in adjacent points decide whether the query point belongs to a discontinuity plane. These methods iteratively expand regions of similar normal until a threshold change is surpassed, indicating the presence of a discontinuity [24, 25]. Being an iterative process, region growing methods demand high computational resources, especially when applied to large point clouds. Additionally, they may produce multiple clusters for the same discontinuity plane or miss out on important ones because of their sensitivity to point normal variability, making result interpretation challenging. In some alternative approaches, researchers have tried to formulate techniques to mitigate the ambiguity in selecting local neighbourhood support regions in point normal estimation for discontinuity mapping. Among those techniques, a popular approach is to use an octree structure to divide the point cloud into variable-sized voxels based on local point density, followed by a plane fitting method [18, 26]. Another technique, utilising Delaunay triangulation to derive a triangular irregular network (TIN) from the point cloud, can be used to estimate surface normals directly from the TIN, thus eliminating the need to define a separate local support region [18, 27]. However, TIN-based methods require additional pre-processing steps, such as surface smoothing filters, to account for variations in point precision and sampling density. In an investigation, [28] described a method for introducing robustness in normal estimation and clustering for discontinuity identification in a noisy point cloud by using local point descriptors to capture unique surface properties. However, this technique of descriptor generation requires extremely high processing times, sometimes over a day, for large-scale point clouds, and hence is unsuitable for real-world applications.

In most cases, the techniques discussed in the literature are applied to point clouds from the open-source RockBench dataset [29]. This dataset contains multiple point clouds of rock-face sections from rock masses of various lithologies, including granite, gneiss, limestone, sandstone, and quartzite. These high-resolution point clouds are generated using photogrammetry and terrestrial laser scanning and are widely used for benchmarking structure-mapping algorithms. However, RockBench represents an idealised scenario where the surfaces are exceptionally clean, well structured, and exhibit minimal surface variability. As a result, many existing techniques are tuned to such ideal data and often lack the robustness required for real-world point clouds, which commonly contain significant noise, granular surface roughness, and irregularities. Furthermore, RockBench datasets are not enclosed. They typically represent single exposed rock faces such as road cuts or isolated tunnel walls, analogous to analysing only one face of a cube. Although the visible surfaces may contain meaningful structures, the data do not represent fully enclosed geometries with orientations spanning the entire spectrum. In contrast, real underground tunnels or mine cavities contain points on at least four of the six sides, forming inherently enclosed datasets with cyclic orientation. Techniques developed for non-enclosed point clouds are therefore not equipped to address the specific challenges introduced by data cyclicity in fully enclosed or partially enclosed underground environments.

In this study, a new approach is proposed for automated characterisation of discontinuity sets in real-world enclosed rock-face point clouds. The approach incorporates a single-shot filtering step that preserves discontinuity planes while robustly eliminating noise and high-curvature artefacts in a single pass using a frequency-domain signal-processing technique. In addition, an innovative cyclic orientation transformation scheme is introduced to resolve orientation-cyclicity ambiguities and enable accurate clustering of orientation data. A hierarchical clustering procedure, followed by a plane filtering step, is then employed to identify well-defined discontinuity planes and sets while accommodating varying orientation density distributions without requiring user-defined cluster numbers. These techniques and methodologies are presented in detail in Section 2. The approach is developed, tuned, and tested on real-world point-cloud scans acquired from underground mine stope cavities. Finally, the results are compared against widely used structure-mapping techniques to demonstrate the robustness and efficiency of the proposed approach, as discussed in Section 3.

## 2. Materials and methods

This section presents a description of the point cloud datasets used in the study, followed by a detailed elaboration of the single-shot data filtering strategy, the cyclic orientation transformation scheme, the hierarchical clustering technique, and other post-processing steps employed for the automatic characterisation of discontinuity sets in enclosed rock faces.

### 2.1. Study Area and Data Collection

For this study, point cloud datasets were required that are enclosed or cyclic in nature, representing a three-dimensional space with exposed rock surfaces on at least four of the six possible sides, rather than the typical flat rock or tunnel surfaces commonly examined in previous studies. Such configurations more accurately represent real-world underground cavities, where rock surface orientations are distributed across the full orientation spectrum. To meet this requirement, 3D point clouds of underground stopes were selected. Stopes are large, excavated chambers in mines created during ore extraction. Owing to their partially enclosed geometry, stopes provide an ideal environment for performing discontinuity set characterisation on enclosed rock faces of stopes. Structure mapping of stopes is critical for ensuring safety, maintaining stability, and improving operational efficiency during ore excavation.

The stope point cloud data used in this study were acquired from a metal mine in Australia, where mining operations extend over an area of approximately 3 km × 6 km and reach depths of around 900 m below sea level. The mine extracts large polymetallic ore bodies using the sublevel open stoping method. Individual stopes are typically excavated in two or more blast sequences, with stopes ranging from 25 m to 100 m. The stope scans are collected by the mine site using a Hovermap scanner, a SLAM-based mobile laser scanning system, which is mounted on a DJI Enterprise M210 drone [30]. The Hovermap uses a Velodyne 16-channel mechanical LiDAR mounted on a 360° rotating head, enabling a complete spherical field of view ideal for capturing complex underground geometries such as stopes. The resulting point clouds are subsequently georeferenced, transforming their local coordinate systems into a global reference frame with true north for accurate orientation estimation. Georeferencing is performed via an affine transformation based on ground control targets whose global 3D coordinates are determined through total station surveys conducted by the mine site. For this study, six stope point clouds have been acquired from this mine site. Figure 1 presents the detailed stope scans along with their respective dimensions. The nominal point spacing and point density of the point clouds are ~2.5 cm and ~1600 points/m$^2$, respectively. For analysis and visualisation of the results in the subsequent sections, Stope 1 was primarily used as the representative case study.

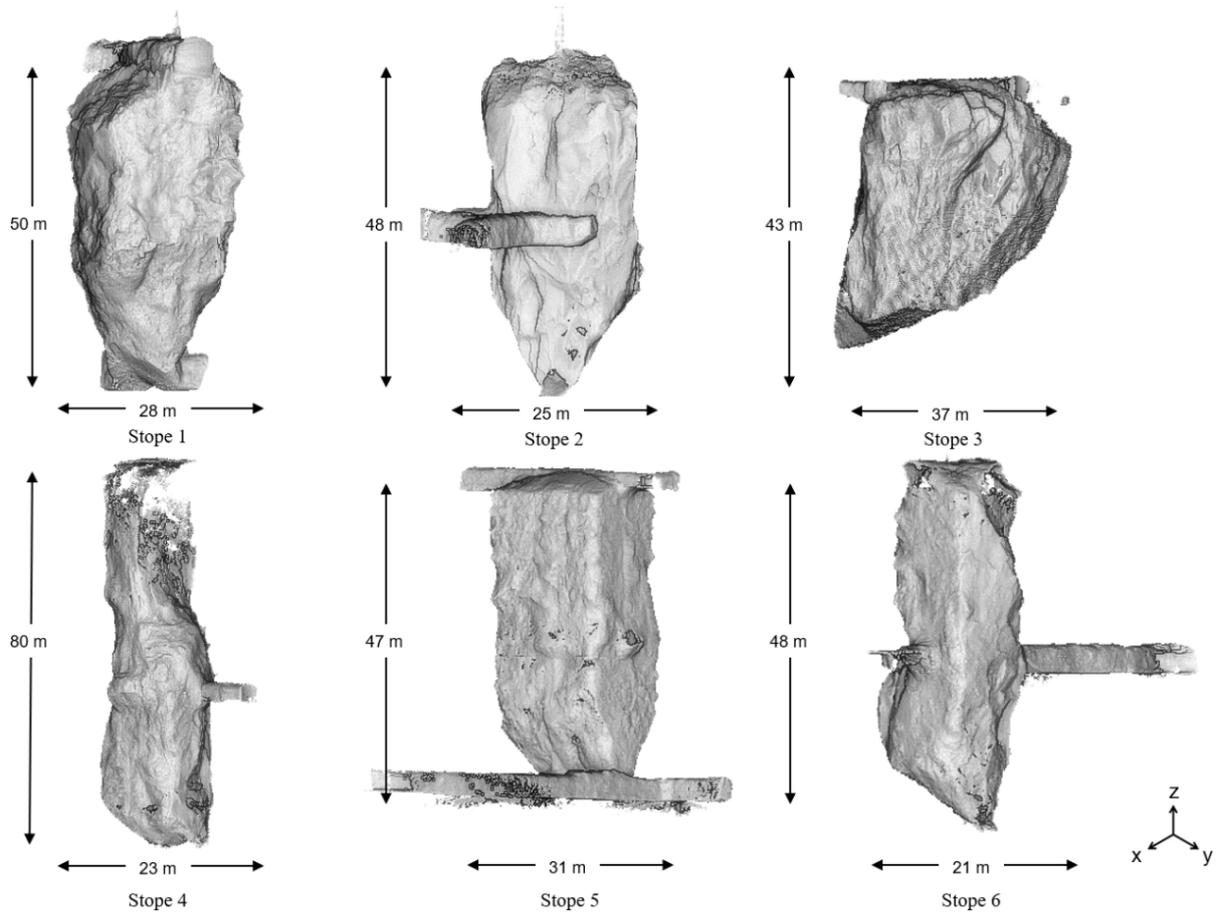

*Figure 1. Stope 3D point clouds acquired from a mine site.*

Additionally, to evaluate the effectiveness of the cyclic orientation transformation scheme proposed in this study, a simulated 3D point cloud of an icosphere with a radius of 10 m was generated, centred at the origin. A low-polygon icosphere was created by subdividing a regular icosahedron once, resulting in a geodesic dome–like structure composed of evenly distributed triangular facets over a spherical surface. The resulting mesh contains 80 triangular faces, with each face having an antiparallel counterpart on the opposite side of the sphere, forming 40 pairs of parallelly oriented faces. This configuration provides an ideal case for assessing cyclicity, as it encompasses the full orientation spectrum with distinct sets of parallelly oriented opposite surfaces. The mesh was subsequently converted into a 3D point cloud through random spatial sampling for use in the analysis. An illustration of the icosphere can be seen in Figure 2.

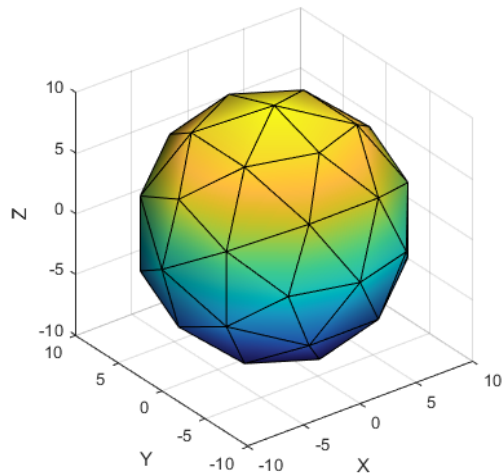

*Figure 2. An illustration of an icosphere created by subdividing a regular icosahedron once.*

## 2.2. Single-Shot Data Filtering

In this study, a single-shot data filtration strategy was adopted. This strategy employs a unique signal processing technique to eliminate non-planar points such as surface variations caused by erroneous points and laser backscattering, data noise, loose gravels, rock fragments, granular high-curvature regions, and edges of planes, while preserving discontinuity planes based on their signal characteristics in the frequency domain [31-34]. This method is highly robust, capable of filtering multiple types of unwanted noise and non-discontinuity plane points in a single step, unlike conventional approaches that rely on a combination of algorithms such as k nearest neighbour noise filters, connected component filters, and range or radius based filters, each of which targets only one type of noise at a time. In addition to its robustness, the computational complexity is significantly reduced, as a single comprehensive filtering method is used instead of multiple sequential ones, efficiently isolating the prominent discontinuity plane regions.

To exploit the unique signal characteristics of points within the point cloud, a spherical support region is defined for generating the corresponding signals. The radius of this region is adaptively determined as a function of the average point spacing (*PS*) of the cloud, as shown in Equation 1 and following the approach adopted in previous studies [28, 35]. The point spacing itself is automatically computed using a 3D Delaunay triangulation of the entire point cloud. The adaptive function reaches its maximum at a *PS* of approximately 0.15 m, and for practical point clouds with sufficient density for structural mapping (*PS* < 0.15 m), it produces an optimally scaled neighbourhood size that adjusts dynamically to the underlying point spacing. In this study, for a nominal point spacing of 0.025 m, the resulting radius of the support region was 0.115 m.

$$Radius\ of\ influence = 5 \times PS - 16 \times PS^2 \tag{1}$$

For this filtering strategy, a signal is generated for every point in the point cloud. For each query point $Q_i$ in the cloud, the elevation angle $E_{ij}$ and azimuth angle $A_{ij}$ are computed for all neighbouring points $P_{ij}$, located within the spherical support region, to the query point $Q_i$ (where *i* denotes the index of the query point and *j* denotes the index of points within the support region) as shown in Equations 2-3. By definition, the elevation angles and the azimuth angles range from -90° to 90° and 0° to 360°, respectively. For each point $Q_i$ lying on a planar surface, the elevation angles plotted against ascending values of azimuth angles closely resembles a sinusoidal signal, as illustrated in Figure 3a. This behaviour does not hold true for points on non-planar surfaces, where the resulting signal deviates substantially from a sinusoidal form, as shown in Figure 3c. These differences can be effectively exploited in the frequency domain. A Fast Fourier Transform (FFT) is applied to convert the signal from the time domain into the frequency domain. The resulting amplitudes of the frequency spectrum show that points on planar surfaces exhibit predominantly linear behaviour with low standard deviation in the secondary frequency components (Figure 3a), whereas points on non-planar surfaces show substantially higher variation in these components (Figure 3c). Through empirical observation across the entire dataset, it was found that points with a standard deviation greater than one in the secondary components are non-planar regions and were therefore removed from the point cloud as noise. The outcome of this filtering strategy for the test stope is shown in Figure 3b, where the retained planar discontinuity points are coloured grey and the filtered non-planar noise points are coloured blue.

$$Elevation\ Angle\ E_{ij} = tan^{-1}\left(P_z / \sqrt{P_x^2 + P_y^2}\right) \tag{2}$$

$$Azimuth\ Angle\ A_{ij} = tan^{-1}\left(P_y / P_x\right) \tag{3}$$

*Where $P_x$, $P_y$, $P_z$ are the 3D Cartesian coordinates of point $P_{ij}$*

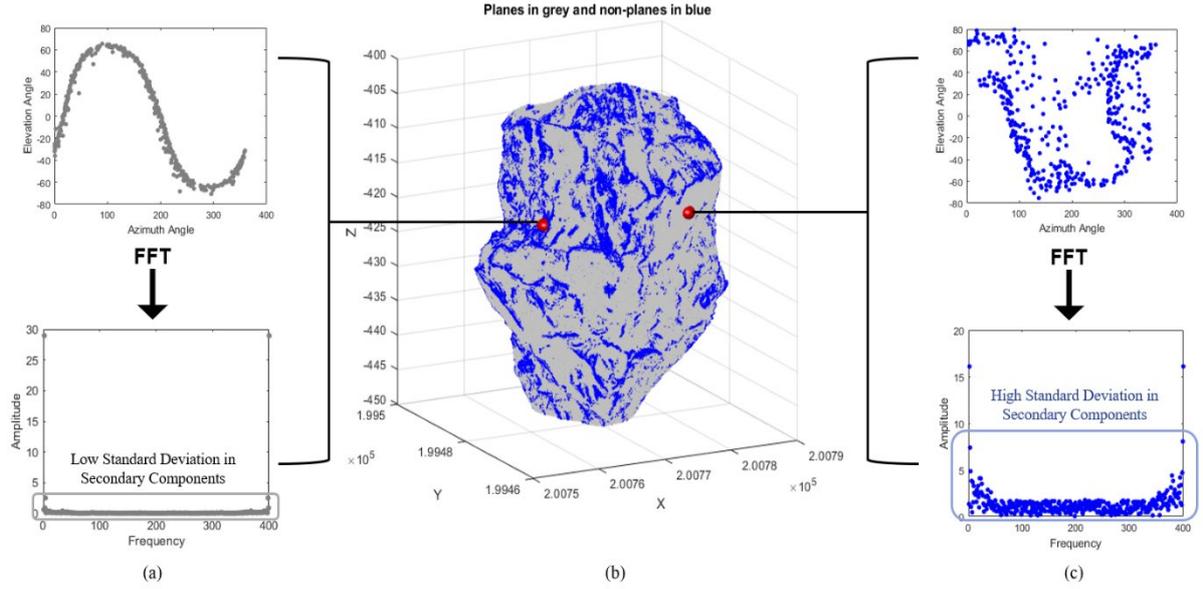

*Figure 3. Single-shot data filtering strategy. (a) Signal characteristics of planes in time and frequency domains. (b) Planes and non-planes colourised based on the standard deviation in amplitude. and (c) Signal characteristics of non-planes in time and frequency domains.*

### 2.3. Cyclic Orientation Transformation

To identify and cluster the discontinuity sets in the filtered point cloud in the next section, the 3D point cloud dataset must first be converted into a corresponding set of point orientations. This is achieved by first estimating a surface or point normal for every point in the cloud. For each point, a principal component analysis (PCA) is performed on the local neighbourhood bounded by the spherical support region defined by the radius in Equation 1. The point normal or unit normal vector ($N$) for each point is then calculated as the eigenvector associated with the smallest eigenvalue of the PCA, representing the direction of least variance and thus the surface normal. Once the point normals are determined, the orientation of each point is calculated and expressed as a pair of dip angle ($DA$) and dip direction ($DD$) values, computed using Equation 4. The dip angle for each point ranges from 0° to 90° and represents the acute angle between the local planar surface and the horizontal plane, measured perpendicular to strike and in the direction of maximum downward inclination. The dip direction ranges from 0° to 360° and denotes the azimuth of the line of maximum downward inclination, measured clockwise from true north.

$$Point\ orientation = \begin{cases} DA = \cos^{-1}(N_z) \\ DD = \tan^{-1}\left(\frac{N_x}{N_y}\right) \\ \begin{cases} DA = 180° - DA \\ DD = DD + 180° \end{cases}, for\ DA > 90° \end{cases}$$

Where $N_x$, $N_y$ and $N_z$ are the vector components of point normal $N$      (4)

Most studies in the literature treat the dip angle–dip direction pair as standard Cartesian coordinates and perform clustering directly in that Cartesian space. However, orientation data are inherently polar and exhibit circular behaviour. This limitation is usually negligible when only a single rock face or tunnel wall is analysed, as the orientations typically occupy a limited portion of the full polar range, reducing the likelihood that cyclic effects introduce noticeable errors. In contrast, for real-world full scans of enclosed underground environments such as stope cavities, point orientations span nearly the entire orientation spectrum, making these cyclic effects much more pronounced during discontinuity set estimation. Although dip angles of 0° and 90° represent the maximum possible angular separation and are depicted accurately in Cartesian space, the same does not hold for dip direction. Dip direction shows true polar cyclicity in which 0° and 360° denote the same physical direction, yet the Cartesian representation places them at opposite ends of the axis, implying maximum separation. When orientation data include values near this cyclic boundary, the resulting discontinuity set estimation in the Cartesian representation can significantly distort clustering outcomes. Therefore, a cyclic orientation transformation is required to correctly represent the geometry and avoid misclassification in discontinuity sets.

To this end, a custom cyclic orientation transformation scheme, loosely inspired by the general projection principles of Schmidt–Lambert equal-area method, is used to generate transformed polar orientations that preserve polar cyclicity in Cartesian space. In this scheme, each *[DA DD]* pair is transformed into *[$D_x$ $D_y$]* using Equations 5–7, providing a Cartesian representation that maintains the inherent cyclic behaviour of the original polar variables. An illustration of this transformation is shown in Figure 4. The transformation of the entire orientation spectrum forms a unit circle in the Cartesian space because each dip angle corresponds to a fixed radial distance from the origin, and varying the dip direction from 0° to 360° sweeps a full rotation around the origin from true north, represented by the positive y-axis in the Cartesian space. As the dip angle increases from 0° to 90°, these radii expand smoothly from zero to one, so the full set of *[DA DD]* orientation values traces concentric circles that collectively fill a unit circle. In doing so, the dip direction naturally wraps in on itself at the 0°–360° boundary, with both values representing the same physical orientation for a given dip angle, thereby preserving the intrinsic cyclicity of the data within the Cartesian domain.

$$radius = \frac{\sin(DA)}{1 + \cos(DA)} \quad (5)$$

$$D_x = radius \times \sin(DD) \quad (6)$$

$$D_y = radius \times \cos(DD) \quad (7)$$

*Where DA, DD are dip angle and dip direction orientations, and $D_x$, $D_y$ are transformed polar coordinates.*

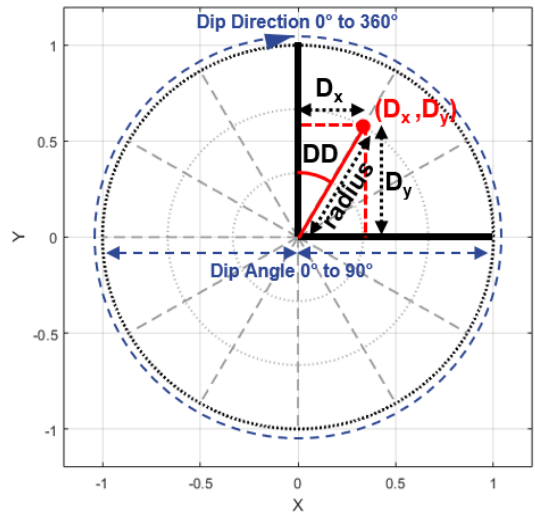

*Figure 4. Cyclic orientation transformation scheme.*

To better understand this phenomenon, a couple of test cases on planar meshes were evaluated, as shown in Figure 5. In the first case, designed to illustrate the circular rotation property of dip direction, twelve planar meshes were simulated with a fixed dip angle of 45° and dip directions spanning the entire 360° range in 30° increments (Figure 5a). When the orientation poles of these planes are calculated and plotted directly as standard Cartesian coordinates *[DA DD]*, the inherent flaw becomes immediately apparent (Figure 5b). Poles of planes 1 and 12 appear at opposite ends of the axis with a separation of 330°, despite being adjacent to each other with only 30° dip direction separation in the actual geometry, as seen in Figure 5a. The cyclic orientation transformation scheme corrects this issue, as shown in Figure 5c. The transformed orientation poles rotate smoothly around the origin, with each adjacent plane rotated by the expected 30° from true north and lying at the same radial distance, reflecting their identical dip angle. This produces a clear circular pattern in the Cartesian space, where poles of planes 1 and 12 are correctly placed 30° apart, preserving the cyclicity of the orientations. Imaginary spherical grid lines are included in the Cartesian plot to better visualise the transformed orientation bounds.

Similarly, in the second test case, the dip angle is progressively varied from 0° to 90° in 15° increments while keeping the dip direction fixed at 90° (Figure 5d). This case is used to illustrate how the dip angle controls the radial distance of the transformed orientation from the origin in Cartesian space. The standard Cartesian plotting of the orientation poles for this case is shown in Figure 5e. More importantly, Figure 5f demonstrates that, for a fixed dip direction, the transformed orientation poles moves smoothly from a radial distance of zero to one as the dip angle increases, indicating that dip angles of 0° and 90° represent the maximum possible separation, which is consistent with the geometric configuration of the planes in Figure 5d. Together, these two test cases clearly show how the cyclic transformation scheme preserves the inherent cyclic behaviour of polar orientations when expressed in Cartesian space. For clarity, the planes and their corresponding orientations in both test cases have been colour-coded according to the legend. A demonstration of the effectiveness of this scheme will be presented in the results section on the simulated icosphere used in this study.

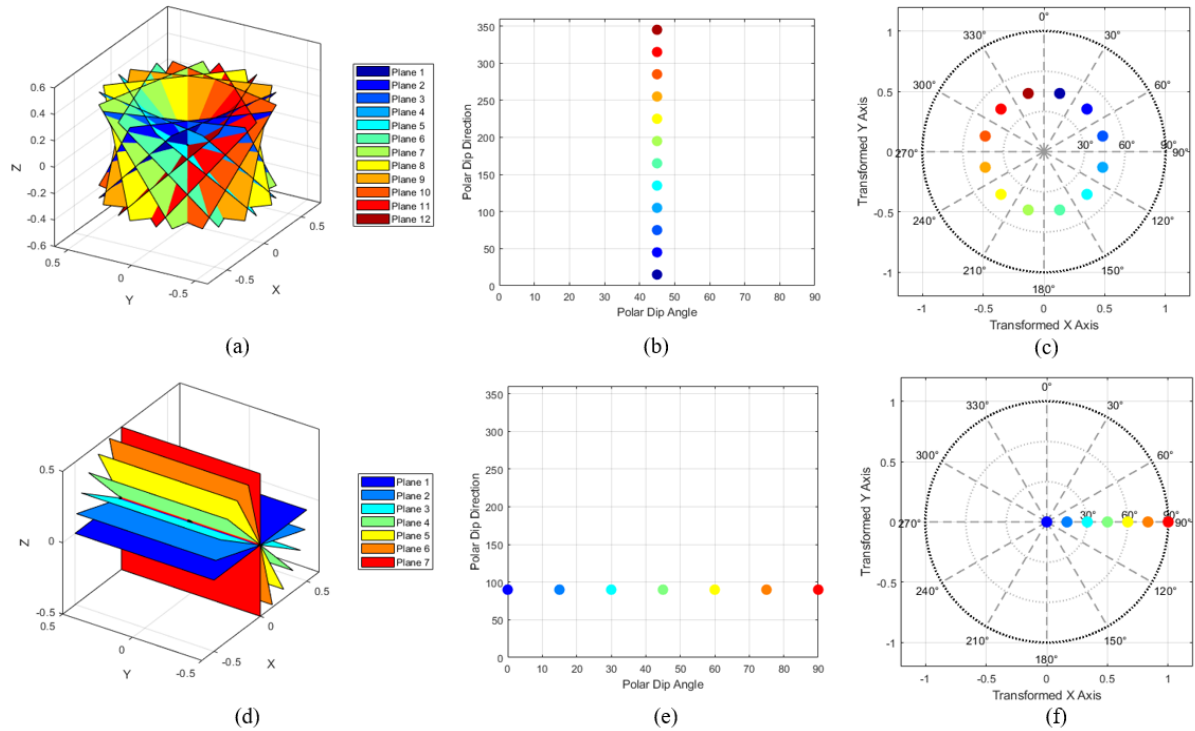

*Figure 5.* (a) Test case 1 - Simulated plane meshes with a fixed dip angle of 45° and dip directions ranging from 0° to 360°. (b) Cartesian-style plot of the polar orientations of planes (variable dip direction). (c) Transformed polar orientations of planes (variable dip directions) plotted in the Cartesian plane. (d) Test case 2 - Simulated plane meshes with a fixed dip direction of 90° and dip angles ranging from 0° to 90°. (e) Cartesian-style plot of the polar orientations of planes (variable dip angles). (f) Transformed polar orientations of planes (variable dip angles) plotted in the Cartesian.

### 2.4. Hierarchical Clustering into Discontinuity Sets

After transforming the orientations, the next step is to cluster these transformed points into discontinuity sets of similar orientations. For this purpose, the hierarchical, density-based, non-parametric clustering algorithm HDBSCAN is employed [36-38]. HDBSCAN is chosen over standard centroid-based parametric methods such as k-means because it does not require the number of clusters to be specified a priori. Instead, the cluster structure is derived directly from the hierarchical dendrogram produced by the algorithm. Parametric techniques typically rely on additional validity indices such as the Silhouette score, Calinski–Harabasz index, or elbow criteria to estimate the optimal number of clusters, which becomes computationally expensive and impractical for the large point clouds analysed in this study [4, 39]. Moreover, compared with flat density-based methods like DBSCAN, HDBSCAN does not assume uniform cluster density and is able to identify clusters of varying density, a behaviour that aligns well with real stope orientation data where discontinuity sets often occur with heterogeneous orientation densities. In addition, like DBSCAN, HDBSCAN explicitly identifies and excludes noise points, ensuring that only genuinely dense and well-supported regions form clusters rather than forcing all data into a cluster assignment. Given these characteristics, HDBSCAN is well-suited for grouping the transformed orientations into discontinuity sets.

HDBSCAN begins by transforming the data into a graph in which distances between points are expressed through the mutual reachability distance shown in Equation 8, a metric that adjusts pairwise distances according to local density and therefore stabilises clustering in regions of varying point density. As opposed to standard Euclidean distance, mutual reachability distance adjusts for local density, preventing sparse areas from being falsely linked and dense areas from dominating the clustering. Using these distances, the algorithm constructs a minimum spanning tree (MST) that captures the most essential connectivity structure in the dataset. The MST connects all points using the minimum total mutual-reachability distance, ensuring that only the most essential and density-consistent links are retained. By progressively removing the longest edges in the MST, HDBSCAN forms a hierarchical clustering tree that reveals how clusters emerge and dissolve across different density levels. This hierarchy is then compressed into a condensed tree, which retains only the branches that persist over substantial ranges of density, effectively filtering out transient or unstable groupings. The final clusters are identified by selecting the branches in the condensed tree with the greatest stability, which corresponds to the amount of support they accumulate across the hierarchy. Two parameters, minimum cluster size and minimum samples, are used in this process. The minimum cluster size controls how big a group of orientation points must be before HDBSCAN is allowed to call it a cluster and is empirically set to 10000, ensuring that only sufficiently large and geologically meaningful discontinuity sets are retained, while avoiding spurious small groupings. The minimum samples parameter defines how many points must exist in a neighbourhood for a point to be considered core and is empirically set to 100, providing a moderate density requirement that helps distinguish true discontinuity sets from noise without being overly restrictive. Together, this configuration allows HDBSCAN to identify orientation clusters in a fully unsupervised manner while accommodating the variable density characteristics typical of real stope orientation data.

$$d_{mreach}(a,b) = \max\{d(a,b), core_k(a), core_k(b)\} \tag{8}$$

Where $a$ & $b$ are two data points in the transformed orientation matrix, $d_{mreach}(a, b)$ is the mutual reachability distance between the two data points, $d(a, b)$ is the standard Euclidean distance between the two data points, $core_k(.)$ is the distance from a data point to its $k^{th}$ nearest neighbour, and $k$ is the minimum samples parameter.

### 2.5. DBSCAN and Plane Filtering

After identifying discontinuity sets of similar orientations, DBSCAN is applied to cluster and segregate the individual discontinuity planes within the points corresponding to each orientation set in the 3D point cloud. This is feasible because planes belonging to the same orientation set are typically separated by spatial gaps, allowing DBSCAN to isolate them based on proximity. The algorithm parameters are chosen according to the point-cloud resolution, where the maximum neighbourhood distance is set to twice the nominal point spacing, on the assumption that adjacent planes will be separated by more than $2 \times PS$; and the minimum number of points required within the spherical neighbourhood is set to 20 (~ $\pi/PS$), empirically determined from point-density considerations. The algorithm also helps remove small clustering inaccuracies that propagate from earlier steps. These arise from local surface variations within the spherical support region during point-normal estimation and appear in the transformed orientation space as narrow strips or small patches of misassigned points, which may be detached from the main cluster or may encroach into neighbouring clusters. DBSCAN's noise-filtering capability effectively eliminates these artefacts, ensuring that only coherent planar regions are retained. In addition, any DBSCAN cluster containing fewer than 100 points is discarded as noise, since clusters of this size are too small to represent geologically meaningful planes. For each retained discontinuity plane cluster, a least-squares plane fit is performed, and its orientation pole is computed from the resulting plane normal using Equation 4. These poles are subsequently plotted on a stereonet, with each plane colour-coded according to the discontinuity set to which it belongs, enabling a clear visual assessment of the results.

### 2.6. Validation and Comparison

The discontinuity sets automatically identified by the proposed approach are compared and validated against three categories of methods: (a) widely adopted structure-mapping algorithms such as k-means clustering and planar region growing; (b) commonly used open-source software tools, including Discontinuity Set Extractor and CloudCompare Facets, and (c) manually interpreted structures obtained using the Virtual Compass tool in CloudCompare. All these methods have been discussed in Section 1. In the k-means clustering approach, the Silhouette coefficient is computed for k values ranging from 2 to 10 to determine the optimal number of clusters in the standard orientation data. The k value with the highest coefficient is selected, and k-means++ is then used to partition the dataset accordingly. In contrast, planar region growing iteratively identifies planar segments directly in the point cloud using a normal-deviation threshold of 6° between adjacent points within a plane and a transmission-error threshold of 20° between neighbouring planes. Once individual planes are identified, their orientations are calculated, and sets are subsequently classified using a k-means approach. In contrast, the open-source software Discontinuity Set Extractor adopts a unique strategy, transforming the orientation data into a continuous probability density function and generating a contour map of the distribution. Peaks in this map represent discontinuity sets, subject to a minimum angular separation of 10° between peaks; each set is then defined as all points falling within a 30° conical neighbourhood around its peak. Another open-source tool, CloudCompare, offers the Facets plugin, which segments the entire point cloud into planar facets by fitting local best-fit planes and then growing and merging them based on angular similarity and spatial proximity. Finally, to validate the results from all the automated approaches, prominent discontinuity planes were manually identified using CloudCompare's Virtual Compass tool. This tool functions analogously to a field clinometer, enabling visual inspection and manual fitting of planes directly on the point cloud so that their orientations can be used as ground truth for assessing clustering accuracy.

## 3. Results and Discussions

### 3.1. Single-Shot Filtering

The result of the single-shot data filtering strategy described in Section 2.2 on the test stope point cloud is shown in Figure 6. The strategy preserves the points lying on planar surfaces that constitute the discontinuity planes within them, as seen in Figure 6a, where the planar regions are highlighted in blue and overlayed on top of the original point cloud. The zoomed-in section of Figure 6a clearly shows how the filtering process retains points on these planar surfaces while discarding high-curvature, non-planar regions and noise. This behaviour is further illustrated in Figure 6b, where all filtered-out points are presented in red. These points include those lying on non-planar surfaces, such as discontinuity edges, loose rock fragments, isolated noise points, stray rock-support bolts, and other high-curvature features that do not represent valid discontinuity planes. This can be clearly observed in the close-up overlay of the filtered result on the original point cloud. From these results, it is evident that the strategy is able to isolate the points of interest, i.e. the planar discontinuity regions, in a single pass while effectively removing non-planar and noisy elements, leaving the point cloud ready for subsequent discontinuity set characterisation.

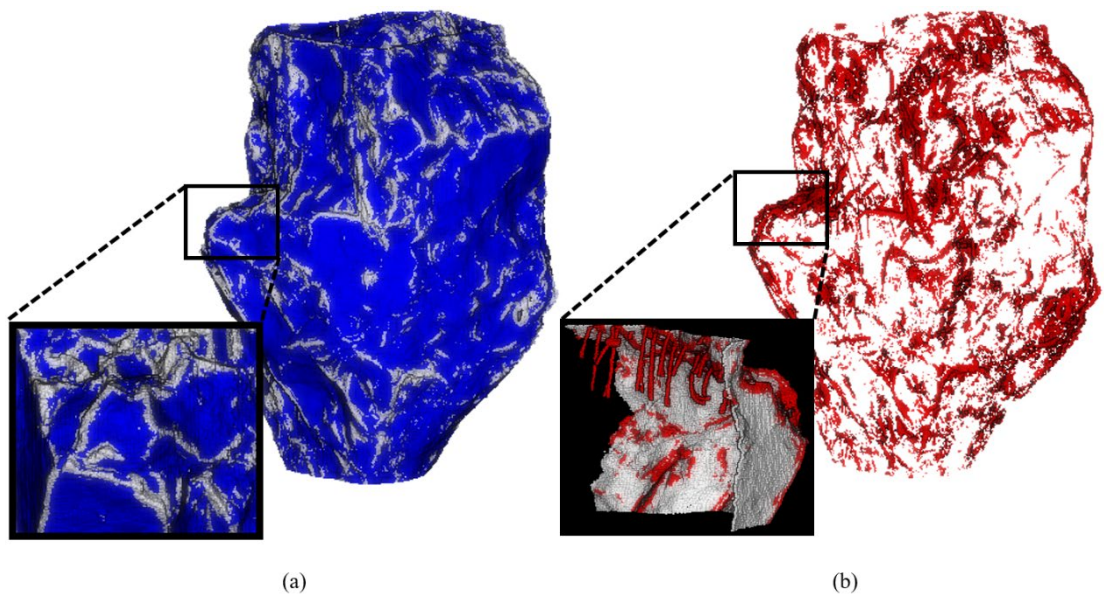

*Figure 6. (a) Resultant filtered points using the single-shot filtering strategy marked in blue and overlayed on top of the original point cloud, with a close-up view of a section for better visualisation of the preserved planar regions. (b) Filtered out points by the single-shot filtering strategy marked in red, with a close-up view of a section for better visualisation of the filtered non-planar regions.*

### 3.2. Icosphere Test

A case study was conducted using the developed method on the simulated icosphere point cloud to evaluate both the cyclic orientation transformation scheme and the hierarchical clustering approach under a fully enclosed geometry. The icosphere provides an ideal test surface, as it represents a completely enclosed spherical form with orientations distributed across the entire spectrum. As shown in Figure 7a, the method successfully identified forty discontinuity sets, fully consistent with the known structure of the icosphere, which contains forty pairs of spatially separated yet parallel triangular faces. The identified planar faces are colour-coded according to the legend, and it is evident that only the true planar facets have been detected, while the non-planar edge regions are correctly excluded and shown in grey. The sets in the point orientations automatically identified through the hierarchical clustering are plotted on a stereonet in Figure 7b. The stereonet clearly displays forty compact, well-separated clusters, each corresponding to one of the forty triangular face pairs, demonstrating the accuracy and reliability of the clustering technique. Notably, the method performs robustly even at the cyclic dip-direction boundary (0°–360°), where conventional Cartesian representation would produce artificial cluster splits. Here, no false clusters are formed at the boundary, confirming that the cyclic orientation transformation behaves correctly in a physical scenario. One such boundary case is highlighted in Figure 7c for plane-pair discontinuity set 9, where the two spatially separated but parallel faces are correctly assigned to a single cluster without any boundary-induced discrepancy. These results demonstrate that the proposed technique is capable of handling fully enclosed rock faces as intended, accurately identifying all orientation sets present in a completely enclosed geometric configuration.

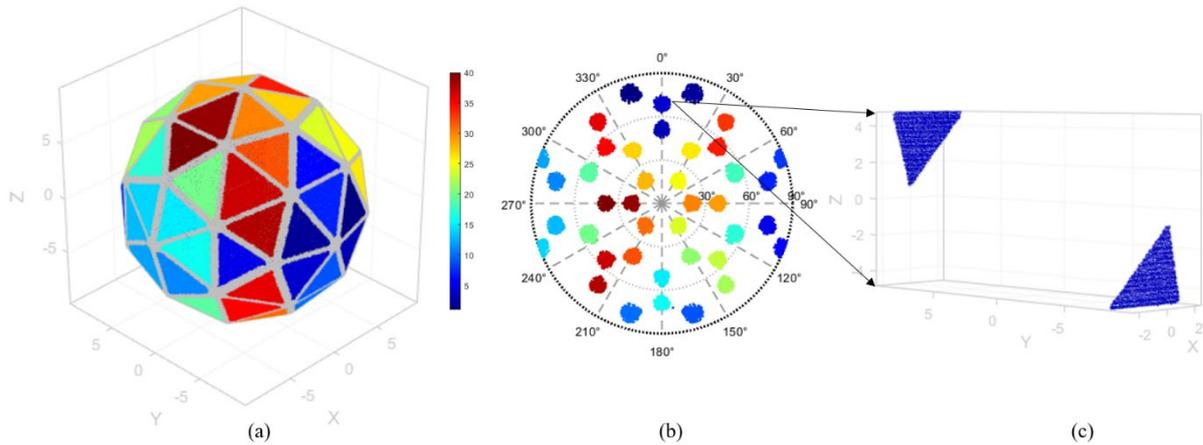

*Figure 7. (a) Identified discontinuity planes marked on the simulated icosphere point cloud and colourised according to the discontinuity set to which they belong. (b) Identified clusters of orientations plotted on a stereonet. (c) Points in the icosphere corresponding to the identified discontinuity set 9. (All sets are colour-coded as per legend)*

**3.3 Clustering Discontinuity Sets**

Hierarchical clustering was applied to the transformed orientation data of the stope point cloud to automatically group them into discontinuity sets of similar orientations. A total of six distinct discontinuity sets were identified. DBSCAN was then used within each set to isolate the individual discontinuity planes, and the corresponding poles were plotted on a stereonet, colourised according to their set membership as shown in Figure 8a. The identified sets exhibit clear separation, due to the inherent noise-handling capability of HDBSCAN, which prevents overlap and suppresses spurious groupings. One of the sets, namely discontinuity set 1, is shown in expanded form in Figure 8b, visualising the corresponding discontinuity planes point cloud and colourised blue according to the assigned set. All planes belonging to this orientation set were successfully detected across all the faces of the stope point cloud. Minor clustering artefacts and plane clusters containing fewer than 100 points were removed by the DBSCAN and plane filtering step and are indicated in black in Figure 8b. The accuracy of the clustering results is further supported by the kernel density estimation (KDE) analysis shown in Figure 8c. The transformed orientation data were used to compute a probability density function, whose iso-density contours were plotted on the stereonet to visualise the dominant peaks in the orientation space. These peaks represent the regions of highest orientation density and therefore the locations where the clustering technique should identify discontinuity sets. The KDE map shows six prominent peaks, and HDBSCAN correctly identifies six corresponding discontinuity sets, with the poles distributed around these high-density regions in a manner consistent with the underlying orientation density. This agreement demonstrates the effectiveness and reliability of the clustering workflow.

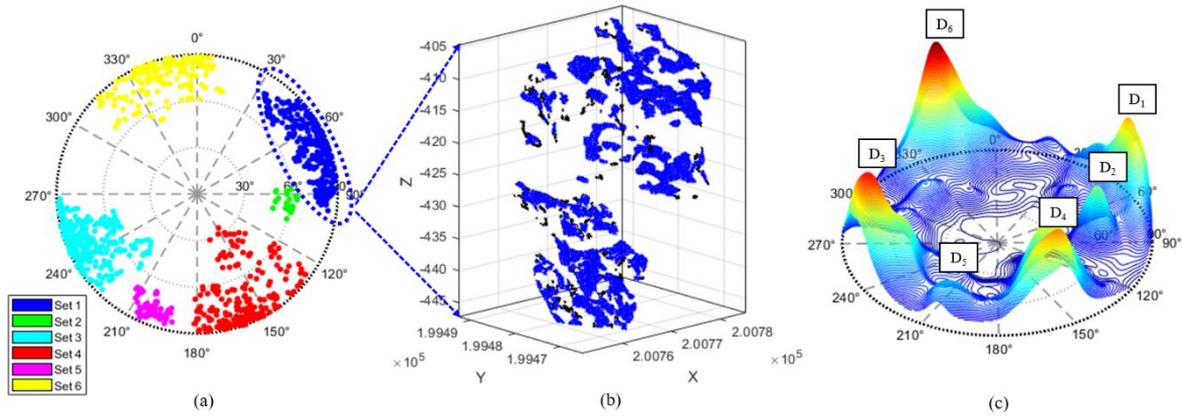

*Figure 8.* (a) Identified discontinuity sets plotted on a stereonet with legend showing set numbers. (b) Expanded view of the point cloud corresponding to the identified poles in discontinuity set 1, with filtered out planes shown in black. (c) Contour map of the point orientations showing peaks in the data to demonstrate the clustering accuracy.

The discontinuity sets and corresponding planes identified in the example stope by the hierarchical clustering are visualised in Figure 9a, with each plane coloured according to its assigned set. While the majority of planes are accurately classified, some minor inaccuracies remain, as visible in the zoomed-in section. These artefacts appear as narrow strips of misclassified points or small encroaching patches that interrupt neighbouring planes. They arise from local surface variations within the point cloud and from tiny spurious segments containing fewer than 100 points, which are considered geologically insignificant and treated as noise. These minor inaccuracies are subsequently removed during the DBSCAN and plane filtering step, producing the final cleaned set of discontinuity planes shown in Figure 9b. The zoomed-in view highlights how this post-processing accurately eliminates residual misclassifications, leaving well-defined discontinuity planes and clear set boundaries while effectively filtering any noise.

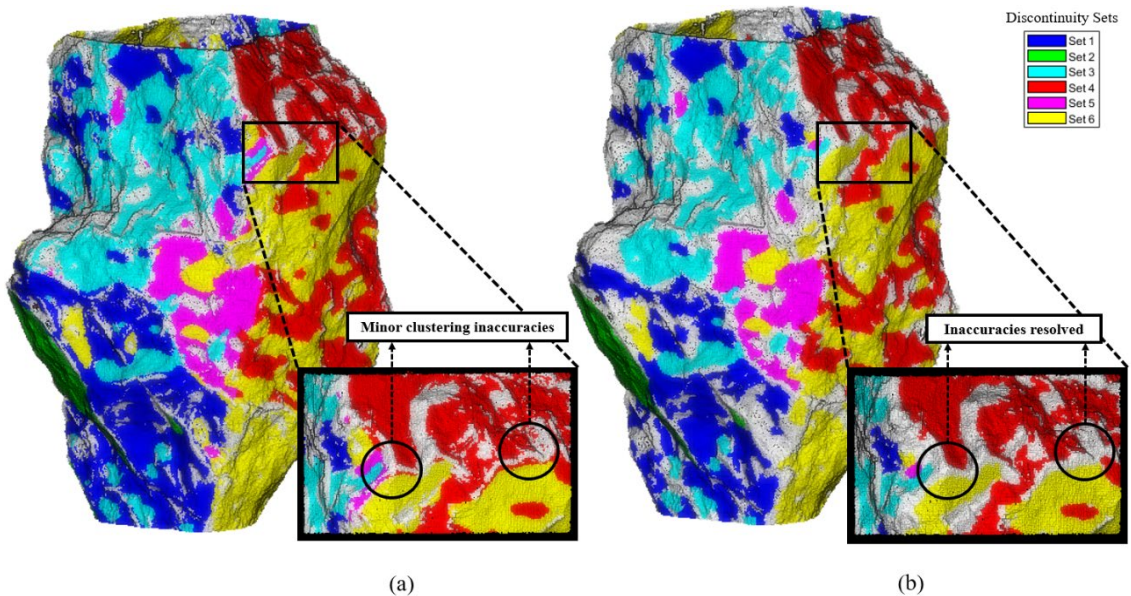

*Figure 9.* (a) Discontinuity sets identified by HDBSCAN visualised in the point cloud with individual sets marked as per legend and a close-up view showing minor clustering inaccuracies. (b) Final discontinuity planes and sets are visualised in the point cloud after DBSCAN and plane filtering, and a close-up view showing that the minor inaccuracies are resolved.

### 3.4. Structure Mapping Comparative Analysis

The structure mapping results of our proposed approach are compared against the widely adopted techniques described in Section 2.6. The identified discontinuity plane poles and sets in the test stope are mapped on the stereonet, as can be seen in Figure 10. The validation set prepared using manual identification of the prominent discontinuity planes shows that there are six distinct discontinuity sets present in the data, each marked with separate colours on the stereonet, as seen in Figure 10f. Although due to inherent limitations in visual perception, the smaller planes could not be identified, the virtual compass mapping can provide an accurate overall representation of the dominant discontinuity sets and serves as a reliable ground-truth reference. The k-means clustering technique is able to identify only four distinct sets in the discontinuity orientations, as shown in Figure 10a. This occurs because the Silhouette index selects four as the optimal number of sets, prioritising maximum separation between clusters and thereby losing the finer and smaller sets. This leads to under-classification, where smaller sets are merged into larger ones. In addition, due to the lack of any noise-handling capability in k-means, there are no clear boundaries between clusters, and the discontinuity poles exhibit significant dispersion around the mean of each set. This characteristic of high dispersion is also evident in the planar region-growing approach shown in Figure 10b. It can also be observed that the number of identified poles is significantly reduced because the strict planarity thresholds only preserve extremely planar regions, which is rarely the case in real-world point clouds with high surface variability, several geologically meaningful planes are therefore lost. Although this approach identifies five sets, one more than k-means, yet one major set still goes undetected due to the aforementioned limitations. The apparent separation between sets is partly an artefact of the reduced pole count, and inter-cluster mixing remains visible, for instance, between the blue and green sets.

The Discontinuity Set Extractor identifies peaks in the probability density function as sets using predefined angular-separation thresholds, as seen in Figure 10c. This causes over-classification because minor peaks cannot be sufficiently suppressed, resulting in a total of ten identified sets. Furthermore, the 30° conical grouping threshold around each peak introduces inter-cluster mixing and over-segmentation in regions affected by surface irregularity. CloudCompare Facets divides the entire point cloud into planar facets using parameters such as maximum angle, maximum relative distance, minimum points per facet, and maximum edge length. The resulting facet orientations are plotted directly on the stereonet in Figure 10d, covering the entire orientation spectrum, making it difficult to distinguish key sets. While the method is useful for viewing plane fragments in 3D, it does not reliably identify the prominent discontinuity sets.

Finally, the discontinuity sets and poles identified by the proposed approach are shown in Figure 10e. The approach identifies and clusters exactly six sets in the data, with the poles concentrated around the mean orientation of each set and exhibiting clear inter-cluster separation. Hierarchical clustering is able to accurately identify all the sets in the data because it can detect clusters even when the underlying data exhibit varying point densities and cluster shapes, which is inherently the case for orientation data. This behaviour, combined with the single-shot filtering strategy and the noise-handling capability of HDBSCAN, produces clean and well-defined clusters. A higher number of poles can also be observed on the stereonet for the proposed approach compared to the other methods, except for CloudCompare Facets, which suffers from extreme over-segmentation, because it successfully identifies even the smaller yet geologically significant discontinuity segments present in the data. Moreover, the yellow discontinuity set crosses the cyclic 0°-360° dip-direction boundary, as can be seen in the ground truth in Figure 10f, and this characteristic is correctly preserved by the proposed approach due to the cyclic orientation transformation scheme employed in this study. In contrast, most other techniques operate in standard Cartesian orientation space and therefore break clusters at this boundary, leading to incorrect cluster assignments for poles near 0° and 360°. Therefore, it can be seen that the proposed approach surpasses the other popular techniques and produces results closest to the ground truth.

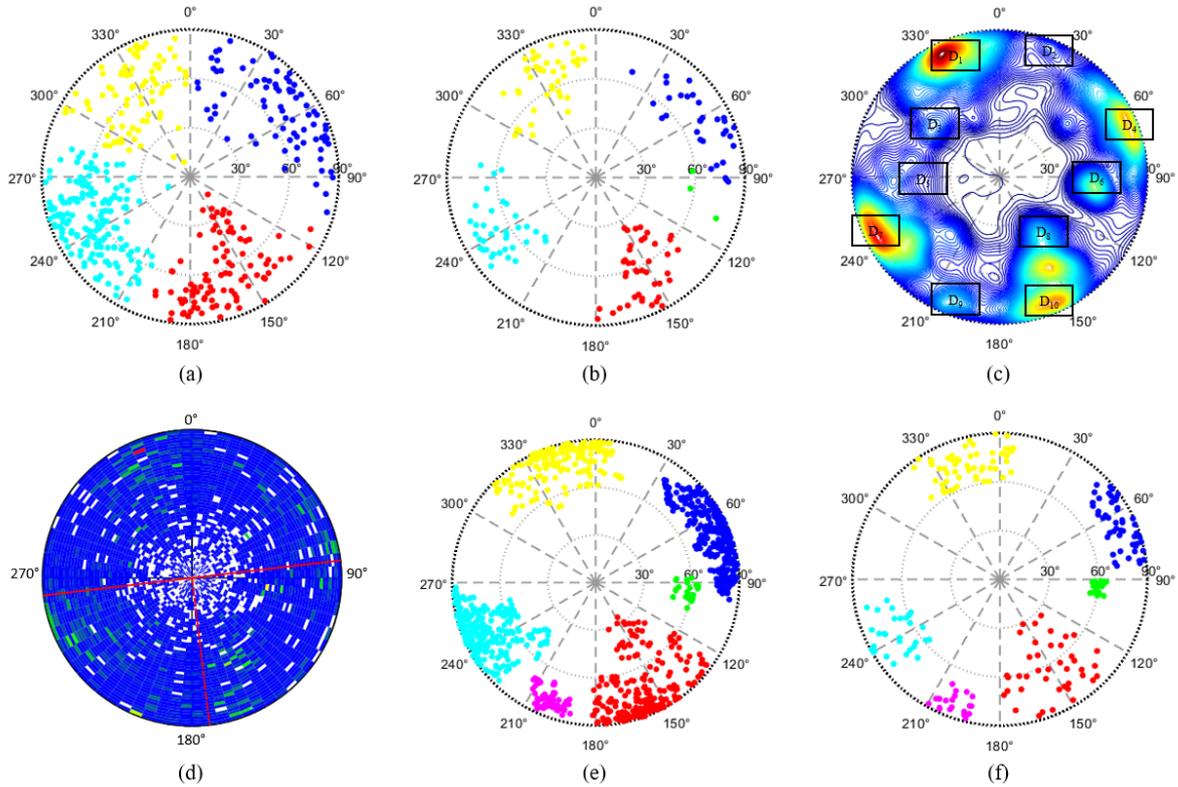

*Figure 10. Automatically identified discontinuity sets by different approaches plotted on a stereonet. (a) k-means clustering, (b) planar region growing, (c) discontinuity set extractor, (d) cloud compare facets, (e) our proposed approach, and (f) validation set prepared using virtual compass.*

The automatically identified discontinuity planes and sets are marked on the 3D point cloud, with legends provided for the individual set colours, as shown in Figure 11 for visual evaluation of the structure-mapping techniques. In Figure 11a, it can be seen that the k-means clustering approach, which suffers from under-classification, produces no clear boundary or separation between the identified discontinuity planes. Moreover, the under-classification leads to several misclassifications, where planes of substantially different orientations are grouped into the same set. In contrast, the planar region-growing approach yields extremely spaced-out discontinuity planes, as shown in Figure 11b. Due to the high surface variation and irregularity present in the point cloud, only very limited, highly planar and smooth regions are identified as planes. As a consequence, many geologically significant discontinuity planes are lost, including one smaller yet important set for which no planes were detected at all, resulting in a loss of critical structural context. This limitation makes planar region growing perform the worst among all the approaches considered.

For the Discontinuity Set Extractor, in addition to over-classification, the threshold-based dispersion of points around density peaks leads to poor separation and mixing between discontinuity planes, as seen in Figure 11c. While the method performs reasonably well in regions with strong planar characteristics, it displays an over-segmentation problem in areas with higher surface variability, where a single continuous plane is fragmented into multiple planes from different sets due to local irregularities. On the other hand, the CloudCompare Facets tool shows the opposite behaviour of planar region growing. Instead of too few planes, it produces an excessively large number. As seen in Figure 11d, the method identifies a total of 7,952 planar facets, and, similar to the Discontinuity Set Extractor, it over-segments the surfaces, resulting in multiple planar fragments for a single discontinuity plane. However, the degree of over-segmentation is considerably higher, leading to ambiguity and extreme clutter in the orientation space.

Finally, the discontinuity planes and sets identified by the proposed approach are shown in Figure 11e. The approach correctly identifies exactly six sets of discontinuity planes in the point cloud, consistent with the ground-truth sets. It achieves a clear balance between plane identification and separation, where all significant planes are detected, with well-defined boundaries between individual planes, unlike the k-means approach. Furthermore, the method is robust in handling regions with high surface variability, where granular high-curvature regions such as loose gravels, rock fragments, bolts, and plane edges are effectively removed from the plane estimation, while the major planar regions are fully preserved, unlike the planar region-growing approach. The discontinuity planes identified by the proposed method show clearer boundaries and more reliable identification in regions of high surface variability, making it substantially superior to the other approaches.

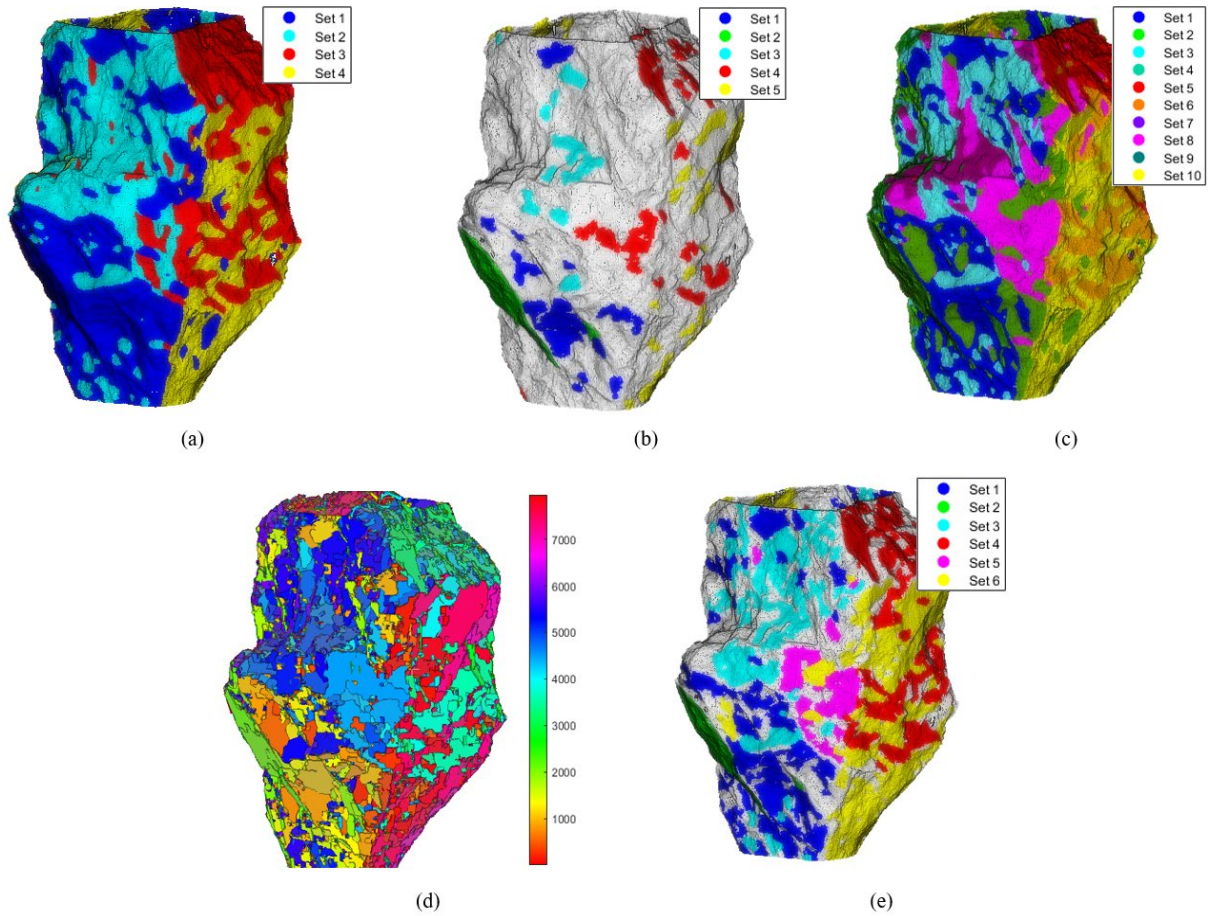

*Figure 11. Automatically identified discontinuity sets by different approaches visualised on the point cloud. (a) k-means clustering, (b) planar region growing, (c) discontinuity set extractor, (d) cloud compare facets, and (e) our proposed approach.*

A comparison of the nominal orientations of the identified discontinuity sets, represented by the mean dip angle $\overline{DA}$ and mean dip direction, along with their respective standard deviations $SD$ in the test stope for each method, is provided in Table 1. In addition, the mean absolute error in the nominal orientations and the dispersion error (defined as the mean absolute error in standard deviation) are computed for all automated techniques relative to the virtual compass results, which serve as the ground-truth reference. The k-means approach performs the worst, yielding orientation errors exceeding ±8° for both dip angle and dip direction, and the highest dispersion errors due to the large spread of poles within each set. All other techniques maintain dispersion errors below ±5.5°, indicating reduced pole scattering. Planar region growing shows modest improvement in orientation accuracy compared to k-means; however, nominal orientation errors still exceed ±5.5°, which remains substantial and unsuitable for reliable structural characterisation. The Discontinuity Set Extractor provides notably better mean-orientation accuracy, with errors below ±4° for both dip angle and dip direction, placing it closer to the ground truth. Nevertheless, its dispersion behaviour is less favourable, which, although improved relative to k-means, performs worse than planar region growing in this metric. This is a direct consequence of its fixed 30° conical grouping threshold, which forces each set to exhibit nearly constant standard deviation values (approximately 10° for both dip angle and dip direction), regardless of the true geometric variability. As a result, dispersion becomes unreliable with artificially low values for larger sets and disproportionately high values for smaller ones. In contrast, the proposed approach outperforms all other methods in both nominal orientation error and dispersion. With mean-orientation errors of approximately ±2° for both dip angle and dip direction, deviations from the ground truth are minimal and fall within the range of subjective variation inherent in manual visual interpretation. The dispersion error remains below ±3°, indicating that the spread of poles within each set closely matches the true structural variability. These results demonstrate that the proposed approach delivers high accuracy, consistent dispersion, and robust performance, making it well-suited for practical application in real-world enclosed point-cloud datasets such as stopes.

*Table 1: A quantitative analysis of identified discontinuity sets by the different automated techniques.*

| Discontinuity Sets | Virtual Compass (*Validation*) | | k-Means Clustering | | Planar Region Growing | | Discontinuity Set Extractor | | Proposed Approach | |
|---|---|---|---|---|---|---|---|---|---|---|
| | $\overline{DA} \pm SD$ | $\overline{DD} \pm SD$ | $\overline{DA} \pm SD$ | $\overline{DD} \pm SD$ | $\overline{DA} \pm SD$ | $\overline{DD} \pm SD$ | $\overline{DA} \pm SD$ | $\overline{DD} \pm SD$ | $\overline{DA} \pm SD$ | $\overline{DD} \pm SD$ |
| $D_1$ | 82.73° ± 4.20° | 68.56° ± 8.53° | 70.62° ± 17.12° | 58.05° ± 19.22° | 77.17° ± 8.44° | 61.38° ± 15.92° | 84.50° ± 10.84° | 71.55° ± 9.73° | 84.73° ± 10.10° | 71.58° ± 12.63° |
| $D_2$ | 60.95° ± 3.39° | 95.42° ± 3.97° | Not Identified | | 65.17° ± 5.42° | 100.83° ± 5.32° | 56.23° ± 11.50° | 90.17° ± 10.16° | 59.95° ± 3.66° | 93.11° ± 4.51° |
| $D_3$ | 79.89° ± 5.35° | 245.74° ± 8.74° | 72.51° ± 12.63° | 240.49° ± 17.13° | 72.57° ± 7.05° | 239.72° ± 11.42° | 82.84° ± 9.88° | 252.48° ± 11.28° | 82.89° ± 8.43° | 246.70° ± 14.64° |
| $D_4$ | 72.47° ± 10.91° | 157.18° ± 10.30° | 60.55° ± 10.81° | 152.98° ± 11.97° | 63.05° ± 13.42° | 150.72° ± 18.57° | 71.88° ± 12.06° | 155.68° ± 12.15° | 71.80° ± 11.78° | 155.17° ± 14.52° |
| $D_5$ | 79.63° ± 3.51° | 200.66° ± 4.71° | Not Identified | | Not Identified | | 84.94° ± 10.29° | 202.91° ± 10.67° | 80.96° ± 3.93° | 202.93° ± 5.21° |
| $D_6$ | 80.25° ± 5.93° | 345.24° ± 7.93° | 63.67° ± 14.74° | 330.67° ± 11.27° | 78.70° ± 10.22° | 335.22° ± 14.91° | 77.89° ± 11.91° | 340.75° ± 9.95° | 83.92° ± 9.43° | 347.89° ± 10.08° |
| **Mean Absolute Error ± Dispersion Error** | | | 12.00° ± 7.28° | 8.63° ± 6.02° | 5.61° ± 2.95° | 7.02° ± 5.33° | 2.95° ± 5.53° | 3.87° ± 3.29° | **1.95° ± 2.34°** | **2.20° ± 2.90°** |

### 3.5. Execution Times

A comparative analysis of the execution times for each structure-mapping approach applied to the test stope can be seen in Table 2. All executions were performed in MATLAB R2023a on a 64-bit Windows system equipped with an Intel W-2245 CPU and 128 GB RAM, except for the CloudCompare Facets method, which operates as a plugin within the CloudCompare software. The test stope contained approximately 4 million points, and the reported times correspond to processing this full dataset. The rate-limiting step for each approach is also listed in Table 2, as these steps largely determine the overall execution time.

*Table 2. Comparison of the execution time by different approaches.*

| Approach | Execution Time (mins) | Rate Limiting Step |
|---|---|---|
| k-Means Clustering | ~ 217 | Silhouette coefficient calculation |
| Planar Region Growing | ~ 18 | Iterative neighbourhood search |
| Discontinuity Set Extractor | ~ 14 | Kernel density estimation |
| CloudCompare Facets | ~ 5 | Unknown |
| Proposed Approach | ~ 8 | Point normal estimation |

For all MATLAB-based implementations (k-means, planar region growing, Discontinuity Set Extractor, and the proposed approach), point-normal estimation is a common processing step and contributes a substantial portion of the computational cost, exhibiting a quasi-linear time complexity. CloudCompare Facets is the fastest method in terms of raw runtime at approximately 5 minutes and identifies planar facets on the point cloud and is used as a plug-and-play software. More importantly, it is unable to identify broader discontinuity orientation sets, so it is not directly comparable as a full structure-mapping solution. The k-means clustering approach is the most computationally expensive among the MATLAB methods. In addition to normal estimation for all points, it requires evaluating the Silhouette coefficient over multiple candidate values of k to select the optimal number of clusters. This step has quadratic complexity, resulting in total runtimes of several hours and rendering the method impractical for large stope point clouds. The planar region-growing technique, by contrast, relies on repeated neighbourhood searches for threshold-based plane expansion. Although still demanding, it is considerably faster than k-means, completing in approximately 18 minutes. For the Discontinuity Set Extractor, beyond the initial normal estimation, the dominant cost arises from accelerated kernel density estimation, with subsequent filtering steps contributing only marginally to the runtime. Its total execution time of roughly 14 minutes is acceptable and usable for mine planning and design scenarios. In the proposed method, the single-shot filtering strategy operates in quasi-linear time in addition to the shared normal-estimation step, as a signal is generated and evaluated for every point to distinguish planar regions from noise. The total runtime is only 8 minutes which is around 40% faster than the next fastest full structure-mapping method studies here (excluding Facets). Combined with its robustness and accuracy, this computational efficiency makes the proposed technique highly suitable for automated structure mapping in real-world, enclosed point-cloud datasets such as stopes.

### 3.6. Structures Across Multiple Stopes

The proposed approach was applied to automatically identify discontinuity sets in all six point clouds described in Section 2.1, collected from the same mine site. The stereonet plot of the identified poles is shown in Figure 12. It can be clearly seen that the proposed technique successfully identifies the discontinuity sets across all stopes, with poles tightly concentrated around their mean orientations and well-separated between sets, demonstrating the robustness and consistency of the method. All stopes except stopes 3 and 5 contain six sets, and these two contain five sets each. The sets are colour-coded according to the legend provided. Because all stopes originate from the same mine site, certain structural sets are expected to persist and propagate across multiple stopes, a trend which is indeed observed in this case study.

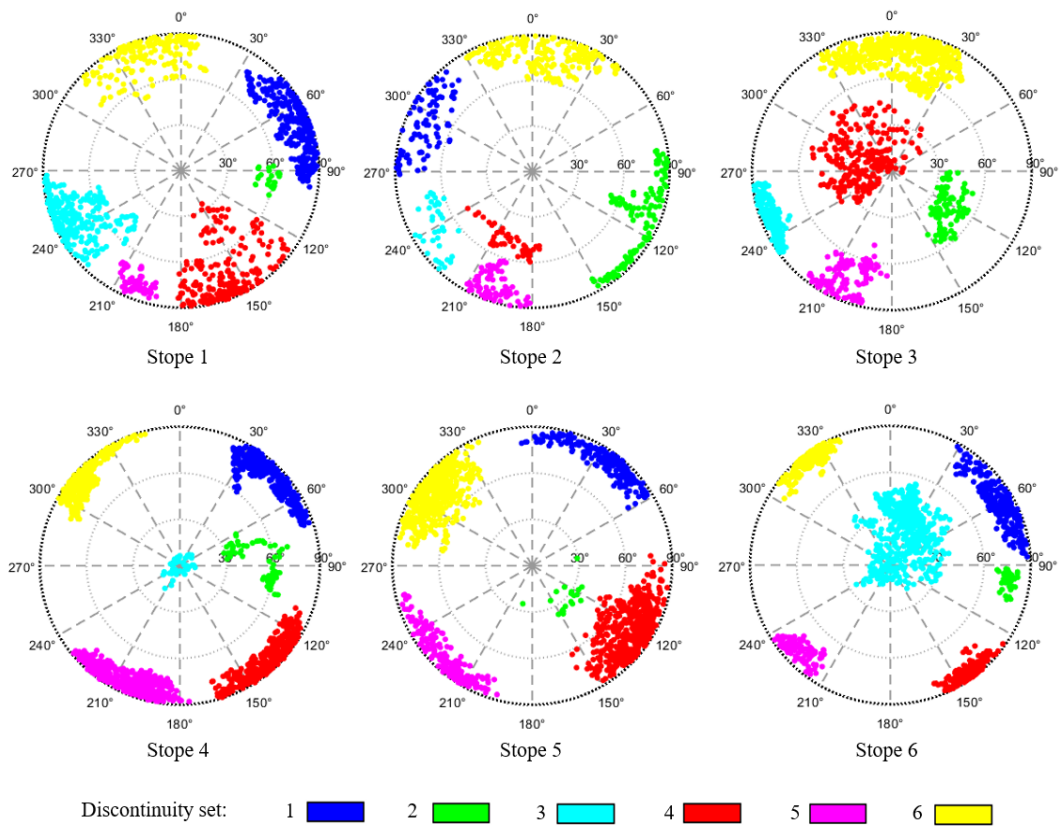

*Figure 12. Identified discontinuity sets in all six stope point clouds acquired from the mine site.*

Sets 2, 5, and 6, coloured green, magenta, and yellow respectively, appear in all six stopes, albeit in different proportions and concentrations of poles, considering a tolerance of approximately ±15° in their nominal dip angles and dip directions. There are inherent differences in the sizes and properties of these sets in different stopes, forming smaller packed clusters in some stopes and more dispersed, bigger clusters in others, due to the differences in the underlying shapes of the clusters in the orientation data. These variations reflect natural differences in the local geometry and the spatial extent of each structure within individual stopes. Similarly, set 1 is consistently present in stopes 1, 4, 5, and 6 with similar orientations, and analogous patterns can be observed for the remaining sets, where structurally equivalent orientations recur across multiple stopes.

Such observations are valuable in mine planning and design, as recognising structural continuity across stopes provides a more holistic understanding of the geological framework and kinematic conditions of the entire mine site. The accuracy and robustness of the proposed structure-mapping approach support these broader interpretations, providing a reliable foundation for structural analysis and aiding the development of future stopes within the site.

## 4. Conclusion

This study presented a comprehensive, fully automated workflow for discontinuity set characterisation in real-world enclosed rock-face point clouds acquired from underground mine environments. By integrating a single-shot filtering strategy, a cyclic orientation transformation scheme, and a hierarchical clustering technique, the proposed approach effectively addressed the challenges posed by noisy, irregular, and fully enclosed geological geometries. The method accurately isolated planar structural features, resolved orientation cyclicity, and automatically identified discontinuity sets, demonstrating strong robustness across real-world stope datasets. Comparative analyses against widely used structure-mapping techniques and ground-truth measurements obtained using the virtual compass further highlighted the superior accuracy, efficiency and substantially reduced computational cost of the proposed approach. These capabilities make the method highly suitable for integration into routine geotechnical workflows, supporting improved rock-mass characterisation, stability assessment, and mine-planning decisions. Future work will focus on extending the workflow to GPU-based implementations to further reduce processing time and approach near real-time characterisation. In addition, further research can investigate deploying such frameworks on edge devices such as onboard computers of UAV LiDAR systems or commercially available LiDAR sensors in modern smartphones and tablets, to enhance underground mine safety and operational efficiency through real-time structural assessments.

## REFERENCES

[1] R. Battulwar, M. Zare-Naghadehi, E. Emami, and J. Sattarvand, "A state-of-the-art review of automated extraction of rock mass discontinuity characteristics using three-dimensional surface models," *Journal of Rock Mechanics and Geotechnical Engineering,* vol. 13, no. 4, pp. 920-936, 2021, doi: 10.1016/j.jrmge.2021.01.008.

[2] H. Daghigh, D. D. Tannant, V. Daghigh, D. D. Lichti, and R. Lindenbergh, "A critical review of discontinuity plane extraction from 3D point cloud data of rock mass surfaces," *Computers & Geosciences,* vol. 169, 2022, doi: 10.1016/j.cageo.2022.105241.

[3] Y. Xing, P. H. S. W. Kulatilake, and L. A. Sandbak, "Effect of rock mass and discontinuity mechanical properties and delayed rock supporting on tunnel stability in an underground mine," *Eng Geol,* vol. 238, pp. 62-75, 2018, doi: 10.1016/j.enggeo.2018.03.010.

[4] C. Pu, J. Zhan, W. Zhang, and J. Peng, "Characterization and clustering of rock discontinuity sets: A review," *Journal of Rock Mechanics and Geotechnical Engineering,* vol. 17, no. 2, pp. 1240-1262, 2025, doi: 10.1016/j.jrmge.2024.03.041.

[5] M. J. Lato and M. Vöge, "Automated mapping of rock discontinuities in 3D lidar and photogrammetry models," *International Journal of Rock Mechanics and Mining Sciences,* vol. 54, pp. 150-158, 2012, doi: 10.1016/j.ijrmms.2012.06.003.

[6] J. Chen, Q. Fang, D. Zhang, and H. Huang, "A critical review of automated extraction of rock mass parameters using 3D point cloud data," *Intelligent Transportation Infrastructure,* vol. 2, 2023, doi: 10.1093/iti/liad005.

[7] S. K. Singh, B. P. Banerjee, and S. Raval, "A review of laser scanning for geological and geotechnical applications in underground mining," *International Journal of Mining Science and Technology,* vol. 33, no. 2, pp. 133-154, 2023/02/01/ 2023, doi: 10.1016/j.ijmst.2022.09.022.

[8] L. Jordá Bordehore, A. Riquelme, M. Cano, and R. Tomás, "Comparing manual and remote sensing field discontinuity collection used in kinematic stability assessment of failed rock slopes," *International Journal of Rock Mechanics and Mining Sciences,* vol. 97, pp. 24-32, 2017, doi: 10.1016/j.ijrmms.2017.06.004.

[9] S. Thiele, L. Grose, and S. Micklethwaite, "Compass: A CloudCompare workflow for digital mapping and structural analysis," presented at the EGU General Assembly Conference Abstracts, Vienna, Austria, April 01, 2018, 2018. [Online]. Available: https://ui.adsabs.harvard.edu/abs/2018EGUGA..20.5548T.

[10] D. Patra, C. Baylis, P. Ranasinghe, B. Banerjee, and S. Raval, "A UAV Laser Scanning Technique for Automated Mapping of In-Stope Structural Discontinuity Sets in Underground Mines," presented at the IGARSS 2025 - 2025 IEEE International Geoscience and Remote Sensing Symposium, 2025. [Online]. Available: https://ieeexplore.ieee.org/stampPDF/getPDF.jsp?tp=&arnumber=11243341&ref=.

[11] B. Hutchison and J. Chambers, "Monitoring of structurally controlled deformations at the Kanmantoo copper mine," presented at the Proceedings of the 2020 International Symposium on Slope Stability in Open Pit Mining and Civil Engineering, 2020.

[12] A. J. Riquelme, A. Abellán, R. Tomás, and M. Jaboyedoff, "A new approach for semi-automatic rock mass joints recognition from 3D point clouds," *Computers & Geosciences,* vol. 68, pp. 38-52, 2014, doi: 10.1016/j.cageo.2014.03.014.

[13] A. Riquelme, M. Cano, R. Tomás, and A. Abellán, "Identification of Rock Slope Discontinuity Sets from Laser Scanner and Photogrammetric Point Clouds: A Comparative Analysis," *Procedia Engineering,* vol. 191, pp. 838-845, 2017, doi: 10.1016/j.proeng.2017.05.251.

[14] A. J. Riquelme, A. Abellán, and R. Tomás, "Discontinuity spacing analysis in rock masses using 3D point clouds," *Eng Geol,* vol. 195, pp. 185-195, 2015, doi: 10.1016/j.enggeo.2015.06.009.

[15] G. Gigli and N. Casagli, "Semi-automatic extraction of rock mass structural data from high resolution LIDAR point clouds," *International Journal of Rock Mechanics and Mining Sciences,* vol. 48, no. 2, pp. 187-198, 2011, doi: 10.1016/j.ijrmms.2010.11.009.

[16] T. J. B. Dewez, D. Girardeau-Montaut, C. Allanic, and J. Rohmer, "Facets : A Cloudcompare Plugin to Extract Geological Planes from Unstructured 3d Point Clouds," *The International Archives of the Photogrammetry, Remote Sensing and Spatial Information Sciences,* vol. XLI-B5, pp. 799-804, 2016, doi: 10.5194/isprs-archives-XLI-B5-799-2016.

[17] M. Vöge, M. J. Lato, and M. S. Diederichs, "Automated rockmass discontinuity mapping from 3-dimensional surface data," *Eng Geol,* vol. 164, pp. 155-162, 2013, doi: 10.1016/j.enggeo.2013.07.008.

[18] J. Chen, H. Huang, M. Zhou, and K. Chaiyasarn, "Towards semi-automatic discontinuity characterization in rock tunnel faces using 3D point clouds," *Eng Geol,* vol. 291, 2021, doi: 10.1016/j.enggeo.2021.106232.

[19] J. Chen, H. Zhu, and X. Li, "Automatic extraction of discontinuity orientation from rock mass surface 3D point cloud," *Computers & Geosciences,* vol. 95, pp. 18-31, 2016, doi: 10.1016/j.cageo.2016.06.015.

[20] S. Slob, B. van Knapen, R. Hack, K. Turner, and J. Kemeny, "Method for Automated Discontinuity Analysis of Rock Slopes with Three-Dimensional Laser Scanning," *Transportation Research Record: Journal of the Transportation Research Board,* vol. 1913, no. 1, pp. 187-194, 2019, doi: 10.1177/0361198105191300118.

[21] J.-w. Zhou, J.-l. Chen, and H.-b. Li, "An optimized fuzzy K-means clustering method for automated rock discontinuities extraction from point clouds," *International Journal of Rock Mechanics and Mining Sciences,* vol. 173, 2024, doi: 10.1016/j.ijrmms.2023.105627.

[22] P. Zhang, J. Li, X. Yang, and H. Zhu, "Semi-automatic extraction of rock discontinuities from point clouds using the ISODATA clustering algorithm and deviation from mean elevation," *International Journal of Rock Mechanics and Mining Sciences,* vol. 110, pp. 76-87, 2018, doi: 10.1016/j.ijrmms.2018.07.009.

[23] D. Kong, F. Wu, and C. Saroglou, "Automatic identification and characterization of discontinuities in rock masses from 3D point clouds," *Eng Geol,* vol. 265, 2020, doi: 10.1016/j.enggeo.2019.105442.

[24] Y. Ge *et al.*, "Automated measurements of discontinuity geometric properties from a 3D-point cloud based on a modified region growing algorithm," *Eng Geol,* vol. 242, pp. 44-54, 2018, doi: 10.1016/j.enggeo.2018.05.007.

[25] X. Wang, L. Zou, X. Shen, Y. Ren, and Y. Qin, "A region-growing approach for automatic outcrop fracture extraction from a three-dimensional point cloud," *Computers & Geosciences,* vol. 99, pp. 100-106, 2017, doi: 10.1016/j.cageo.2016.11.002.

[26] J. Guo, S. Liu, P. Zhang, L. Wu, W. Zhou, and Y. Yu, "Towards semi-automatic rock mass discontinuity orientation and set analysis from 3D point clouds," *Computers & Geosciences,* vol. 103, pp. 164-172, 2017, doi: 10.1016/j.cageo.2017.03.017.

[27] J. Lim and P. Pilesjö, "Triangular Irregular Network (TIN) Models," *Geographic Information Science & Technology Body of Knowledge,* vol. 2022, no. Q2, 2022, doi: 10.22224/gistbok/2022.2.7.

[28] S. K. Singh, S. Raval, and B. P. Banerjee, "Automated structural discontinuity mapping in a rock face occluded by vegetation using mobile laser scanning," *Eng Geol,* vol. 285, 2021, doi: 10.1016/j.enggeo.2021.106040.

[29] M. Lato, J. Kemeny, R. M. Harrap, and G. Bevan, "Rock bench: Establishing a common repository and standards for assessing rockmass characteristics using LiDAR and photogrammetry," *Computers & Geosciences,* vol. 50, pp. 106-114, 2013, doi: 10.1016/j.cageo.2012.06.014.

[30] E. Jones, J. Sofonia, C. Canales Cardenas, S. Hrabar, and F. Kendoul, "Advances and applications for automated drones in underground mining operations," presented at the Proceedings of the Ninth International Conference on Deep and High Stress Mining, 2019.

[31] R. P. Devanna, M. Torres-Torriti, K. Sacilik, N. Cetin, and F. Auat Cheein, "Exploring the Frequency Domain Point Cloud Processing for Localisation Purposes in Arboreal Environments," *Algorithms,* vol. 18, no. 8, 2025, doi: 10.3390/a18080522.

[32] S. K. Singh, B. P. Banerjee, M. J. Lato, C. Sammut, and S. Raval, "Automated rock mass discontinuity set characterisation using amplitude and phase decomposition of point cloud data," *International Journal of Rock Mechanics and Mining Sciences,* vol. 152, p. 105072, 2022/04/01/ 2022, doi: https://doi.org/10.1016/j.ijrmms.2022.105072.

[33] C. Zhang, F. Da, and S. Gai, "Point clouds feature frequency domain analysis based on multilayer perceptron," *The Visual Computer,* vol. 41, no. 2, pp. 1007-1020, 2024, doi: 10.1007/s00371-024-03380-9.

[34] S. Zhang, H. Wang, J.-g. Gao, and C.-q. Xing, "Frequency domain point cloud registration based on the Fourier transform," *Journal of Visual Communication and Image Representation,* vol. 61, pp. 170-177, 2019, doi: 10.1016/j.jvcir.2019.03.005.

[35] D. Patra, P. Ranasinghe, B. Banerjee, and S. Raval, "A Deep Learning Approach to Identify Rock Bolts in Complex 3D Point Clouds of Underground Mines Captured Using Mobile Laser Scanners," *Remote Sensing,* vol. 17, no. 15, 2025, doi: 10.3390/rs17152701.

[36] C. Malzer and M. Baum, "A Hybrid Approach To Hierarchical Density-based Cluster Selection," presented at the 2020 IEEE International Conference on Multisensor Fusion and Integration for Intelligent Systems (MFI), 2020.

[37] R. J. G. B. Campello, D. Moulavi, A. Zimek, and J. Sander, "Hierarchical Density Estimates for Data Clustering, Visualization, and Outlier Detection," *ACM Transactions on Knowledge Discovery from Data,* vol. 10, no. 1, pp. 1-51, 2015, doi: 10.1145/2733381.

[38] L. McInnes, J. Healy, and S. Astels, "hdbscan: Hierarchical density based clustering," *The Journal of Open Source Software,* vol. 2, no. 11, 2017, doi: 10.21105/joss.00205.

[39] A. M. Ikotun, F. Habyarimana, and A. E. Ezugwu, "Cluster validity indices for automatic clustering: A comprehensive review," *Heliyon,* vol. 11, no. 2, p. e41953, Jan 30 2025, doi: 10.1016/j.heliyon.2025.e41953.